\PassOptionsToPackage{numbers,sort&compress}{natbib}
\documentclass{article}
\usepackage{graphicx}
\usepackage[final,dblblindworkshop]{neurips_2026}
\workshoptitle{ATTRIB}
\usepackage[utf8]{inputenc}
\usepackage[T1]{fontenc}
\usepackage{hyperref}
\usepackage{url}
\usepackage{booktabs}
\usepackage{amsmath,amssymb,amsfonts}
\usepackage{nicefrac}
\usepackage{xcolor}
\usepackage{float}
\usepackage{marvosym}
\usepackage{microtype}
\usepackage{nicefrac}       

\newcommand{\dd}{\Delta d}
\newcommand{\infpkg}{pyinfluence}
\newcommand{\matchpkg}{cohortmatch}
\newcommand\blfootnote[1]{\begingroup\renewcommand\thefootnote{}\footnote{#1}\addtocounter{footnote}{-1}\endgroup}

\hypersetup{
  colorlinks=true,
  linkcolor={blue!60!black},
  citecolor={blue!60!black},
  urlcolor={blue!60!black},
}

\title{Attributing Cohen's d: Training Data Attribution for Disease-Related Effects in Normative Age Biomarkers}

\author{
  Jakob Snel\textsuperscript{1,2}, Marc-Andr\'{e} Schulz\textsuperscript{1,2,3,4\,\Letter}\\\\
  \small \textsuperscript{1}Hertie Institute for AI in Brain Health, University of T\"{u}bingen, T\"{u}bingen, Germany \\
  \small \textsuperscript{2}T\"{u}bingen AI Center, University of T\"{u}bingen, T\"{u}bingen, Germany \\
  \small \textsuperscript{3}Department of Psychiatry and Neurosciences, Charit\'{e} -- Universit\"{a}tsmedizin Berlin, Berlin, Germany \\
  \small \textsuperscript{4}German Center for Mental Health (DZPG), partner site Tübingen, Germany
}

\begin{document}
\maketitle

\begin{abstract}
Normative age models are trained to predict chronological age in a nominally healthy cohort. In patients, the gap between predicted and chronological age is read as disease risk.
We attribute the disease-related effect size of the age gap directly to individual training samples, rather than using a prediction-level loss as the attribution target.
For Cohen's $d$, the resulting closed-form influence functional, validated against leave-one-out retraining, ranks training samples by their effect on held-out case-control separation. 
Across four diseases and two biomarker modalities in UK Biobank, removing the 10\% most influential training samples raises held-out disease-related effect size in every seed. It more than doubles the metabolomic-age effect for type~2 diabetes and raises the brain-age effect for multiple sclerosis by roughly a third. Random removal leaves effect size flat even at 50\% removal, confirming the gain comes from which samples are removed, not how many. Flagged subjects, missed by diagnosis-based exclusion, show elevated cardiometabolic markers the model never sees, consistent with subclinical contamination of the reference cohort. For type~2 diabetes, where the method gains most, the marker recovered is HbA1c, the standard measure of blood sugar control. We release our influence-function package \href{https://github.com/maschulz/pyinfluence}{\infpkg{}}.
\end{abstract}

\blfootnote{\Letter~Corresponding author: \texttt{marc-andre.schulz@uni-tuebingen.de}}

\section{Introduction}
\begin{figure}[H]
\centering
\includegraphics[width=1\linewidth]{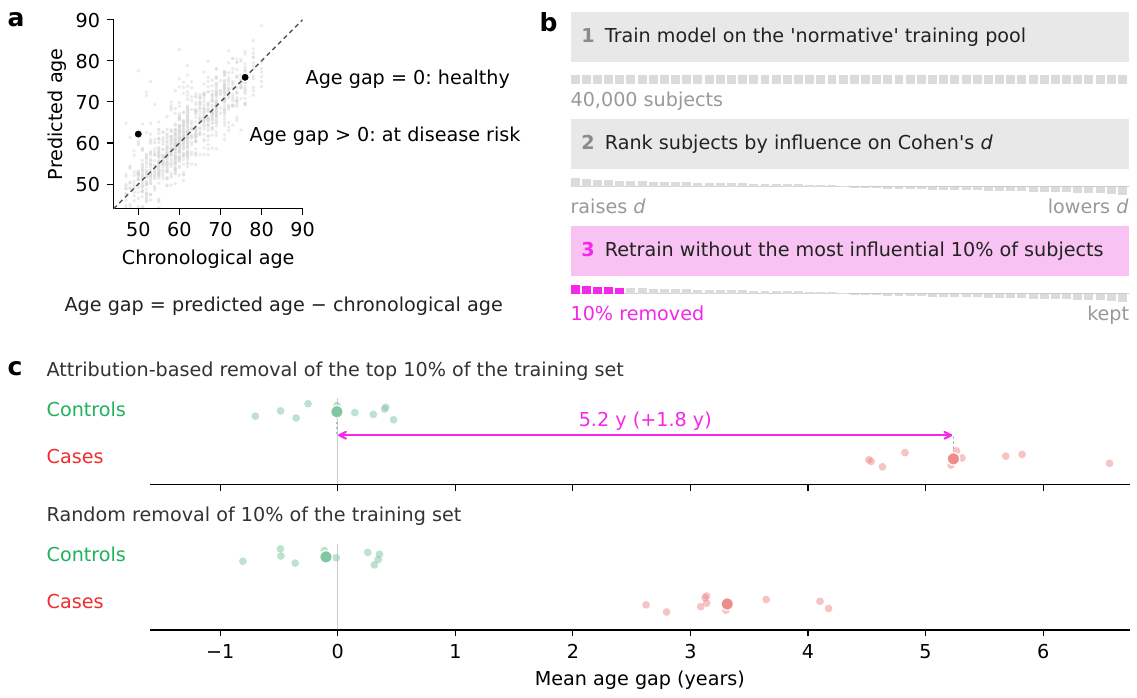}
\caption{Influence-guided removal in the
multiple sclerosis cohort widens the separation in mean age gap between held-out controls and cases.
\textbf{(a)} We plot predicted age against chronological age
for individual subjects in this cohort, and define the age gap as
predicted age minus chronological age. A subject on the diagonal has an
age gap of zero and reads as healthy. The marked subject above it has a
positive age gap and reads as at elevated disease risk.
\textbf{(b)} We train a model
on its nominally healthy training pool of 40{,}000 subjects, rank every training subject by its influence on Cohen's $d$, and retrain the model without the
10\% most influential subjects (magenta), keeping the remainder.
\textbf{(c)} Attribution-based top-10\%~removal (top) widens the
separation in mean age gap between cases and controls to 5.2 years,
a paired gain of 1.8 years over random removal of the same share
(bottom), which leaves a separation of 3.4 years. Small, lighter dots show
single-seed means. The larger, more saturated dot shows the mean across
seeds. Removing the flagged 10\% moves the case group away from its
matched controls while the control group stays near zero.}
\label{fig:intro}
\end{figure}

A normative age biomarker is a model trained to predict chronological age in a nominally healthy cohort \citep{cole2017}. Applied to a patient, it misses, and the gap between predicted and chronological age is read as disease risk (Figure~\ref{fig:intro}a). The gap carries signal because the model never saw disease in training.

Reported brain-age effect sizes for the
same disease vary widely across studies with similar methods
\citep{constantinides2022,jirsaraie2023}. Reference-cohort
contamination is one suspected source \citep{schulz2025}. When a nominally healthy training cohort contains people with pre-diagnostic disease, the
contamination inflates the variance of disease-carrying features, and an optimiser minimising age-prediction error learns to downweight
exactly those features. In fact, models with worse age-prediction
accuracy can show larger disease-related effects \citep{schulz2025}. A control who is pre-diabetic but undiagnosed, for example, still enters training
labelled healthy, and the model learns to treat their elevated glucose-linked features as normal variation, not signal.

\vspace{-10pt}
As in many other domains \citep{oh2023exploring}, current practice for
composing the reference cohort relies on domain experience
\citep{brain_age_paradigm_cole2, estimating_healthy_population, learning_patterns}. Experts build exclusion lists that flag subjects who deviate from the normative ageing trajectory, based on recorded
diagnoses such as diabetes or chronic kidney disease. These lists are
applied a priori, before training, and no post-hoc approach exists to
catch subclinical or undiagnosed cases that never appear on any
exclusion list.

Training data attribution offers a natural way to address this gap. It conventionally identifies the training points most responsible for an individual prediction, for example to debug models or detect dataset errors \citep{koh2017,ghorbani2019,kwon2022,pruthi2020,ilyas2022,park2023}. However, a normative age biomarker is not evaluated by individual prediction accuracy. It is trained on the reference cohort for prediction accuracy, but evaluated by the disease-related effect size of the resulting age gaps, quantified by Cohen's $d$ between held-out cases and matched controls. This mismatch is not unique to our setting.

\citet{nguyen2024towards} find that practitioners' attribution target of interest
varies across users, with no consistent preference for loss. Targets beyond the loss already appear in fairness work, where \citet{wang2022understanding} and \citet{li2022achieving} prune or reweight training samples by their influence on group-level fairness metrics.

We carry this idea to clinical effect size, the case-control separation a biomarker is evaluated by, and call the resulting method Cohen's $d$ influence functions (Figure~\ref{fig:intro}b). This directly targets training samples that reduce the downstream effect size. Under the hypothesis that reference-cohort contamination limits this effect, three predictions follow. First, flagged samples should carry subclinical disease-related phenotypes. Second, removing them should increase evaluation effect size. Third, removing random individuals should not.

Our experiments show all three predictions hold (Section~\ref{sec:effect}). Figure~\ref{fig:intro}c shows the resulting age gap for multiple sclerosis cases and controls. We make the following contributions.
First, \textbf{Cohen's $d$
influence functions}, a closed-form method for attributing disease-related
effect size rather than the training loss, benchmarked against
leave-one-out retraining (Section~\ref{sec:method}). Second,
\textbf{empirical validation across four conditions}. Attribution-based
removal raises disease-related effect size where random removal of the same size
does not, across four diseases and two UK Biobank modalities, in every
seed (Section~\ref{sec:effect}). Third, \textbf{phenotypic characterisation of flagged
samples}. Flagged samples, missed by diagnosis-based exclusion, show elevated cardiometabolic markers, consistent with subclinical contamination of the reference cohort (Section~\ref{sec:pheno}).

\section{Attributing a disease-related effect size}
\label{sec:method}

This section formalises our age-gap biomarker, derives Cohen's $d$ influence functions in closed form, and validates them against leave-one-out retraining.

\paragraph{Biomarker.} For one modality and disease, a ridge regression \citep{perf_reserves} predicts age from features (1{,}440 T1 imaging phenotypes or 251 nuclear magnetic resonance (NMR)
metabolomic analytes) in a training pool of nominally healthy people. We fit a linear age-bias correction $(a, b)$ on training predictions and invert it on test predictions \citep{estimation_smith}. The age gap of subject $i$ is
$g_i = (\hat y_i - b)/a - y_i$. The disease-related effect size is Cohen's $d$ of $g$ between prevalent cases and 1:1 age- and sex-matched controls.

\paragraph{Estimand and functional.} Let $F(D)$ be the effect size we obtain by training the model on the reference cohort $D$ and evaluating it on a fixed, held-out set of matched cases and controls. For every training sample $z_j$, we want the
change $F(D \setminus \{z_j\}) - F(D)$. For ridge, influence on the parameters is closed form \citep{koh2017}. We propagate it through the full pipeline by writing $F$ as an explicit functional of the model's raw predictions $\hat y = X\theta$, where $\theta$ denotes the model's parameters and $X$ is the held-out set's feature matrix, with the bias-correction constants closed over:
\begin{equation}
\label{eq:cohend}
F(\hat y) \;=\; \frac{\bar g_{\text{case}} - \bar g_{\text{ctrl}}}{s_{\text{pooled}}(g)},
\qquad g_i = \frac{\hat y_i - b}{a} - y_i .
\end{equation}
Removing $z_j$ perturbs the fitted parameters $\hat\theta$ by the
closed-form influence step, and the induced change in $F$ is, up to a
positive factor,
\begin{equation}
\label{eq:score}
\mathrm{score}_j \;\approx\; \nabla_{\hat y} F^{\top}\, X\,
  (H + \delta I)^{-1}\, \nabla_\theta \ell(z_j; \hat\theta),
\end{equation}
where $H$ is the Hessian of the training loss, $\delta = 10^{-5}$ stabilises the inverse, and
$\nabla_{\hat y} F$ is the analytic gradient of Equation~\eqref{eq:cohend}. We use only the ranking of the scores, so the omitted positive factor is irrelevant. We compute scores with \infpkg{} \citep{pyinfluence}.

$F$ takes $\hat y$ as input but is Cohen's $d$ of the gap $g(\hat y)$, not of $\hat y$ itself.
Because $g_i$ subtracts each subject's own age,
$\operatorname{var}(g) \neq \operatorname{var}(\hat y)/a^2$, and Cohen's
$d$ computed on raw predictions is a different quantity with a different
gradient. We hold the correction coefficients $(a, b)$ fixed at their full-data values during scoring and do not differentiate their dependence on the removed sample. We refit the coefficients at every actual removal, so any error from holding them fixed shows up as a gap between predicted and realised change, which the removal curves measure. We remove samples in descending score order, the largest estimated $F(D \setminus \{z_j\}) - F(D)$ first.

\paragraph{Selection/evaluation split.} We split all held-out matched cases and controls 50/50 by matched pair into a \emph{selection half}, used only to compute attribution scores, and an \emph{evaluation half}, serving as the test set, used only to report effect sizes. Selection-half effect sizes inflate mechanically with removal. The gap between the two is the overfitting diagnostic, not a result.

\paragraph{Validation.} Exact leave-one-out retraining is tractable for ridge and provides a benchmark for every training sample. At the 10\% cut, the influence and leave-one-out rankings select nearly the same samples. Across all condition-seed combinations, they disagree on fewer than 50 of roughly 4{,}000 removed samples, a set overlap of 98.9\% to 99.8\%. Retraining under each ranking changes evaluation-half Cohen's $d$ by a mean of 0.001 and a maximum of 0.009 across all condition, seed, and removal-level combinations. Influence estimates for nonlinear models can be fragile \citep{basu2021}. Ridge lets us check the closed form against exact retraining.

\section{Experimental setup}
\label{sec:setup}

We use UK Biobank \citep{sudlow2015}, a population cohort of roughly 500{,}000 UK adults with genetic, imaging, and health-record data, across two modalities and four diseases. Brain age from T1 imaging phenotypes at the imaging visit is evaluated against multiple sclerosis (84 evaluation cases) and cerebrovascular disease (404). Metabolomic age from NMR at baseline is evaluated against type~2 diabetes (1{,}000) and chronic kidney disease (1{,}000). Healthy controls matched to diseased cases for evaluation are pure controls, people with no record of the
diagnosis at any time. We exclude individuals diagnosed \emph{after}
the visit from both the training pool and the control pool, applying
best current practice before attribution runs. Any disease-relevant health burden identified afterward is therefore subclinical or undiagnosed, not
something diagnosis-based exclusion already catches. We subsample
training pools to 40{,}000. We choose the ridge penalty once by
cross-validation on the untouched pool and freeze it for every refit.

We compute the ranking once per seed and cut it at 5\%, 10\%, 25\%, and 50\% removal, refitting the model, including the bias correction, at each level and re-scoring the unchanged evaluation half. A control arm removes the same number of samples with a random ranking, holding seed, matched set, split, penalty, and evaluation half identical. We repeat this across 10 seeds per condition.

For each removal level, we compute the paired change $\dd$, the
attribution arm's Cohen's $d$ minus the random arm's, within each
seed, and report its mean and standard deviation across seeds. The absolute effect size carries cohort-draw
variance that is common to every arm within a seed and cancels under pairing.
At 84 evaluation cases the across-seed spread of the absolute $d$ is
$\pm 0.15$, the paired spread $\pm 0.03$ at 10\%
removal, growing to $\pm 0.06$ by 50\%. We run two attribution targets for every cell: the Cohen's $d$ estimand ($d$) and its unnormalised surrogate, the mean age gap. Appendix~\ref{app:setup} contains full reproduction and compute details.

\section{Results}
\label{sec:results}
\subsection{Removing what attribution flags works and removing at random does not}
\label{sec:effect}
\begin{table}[!b]
\centering
\caption{Evaluation-half Cohen's $d$ under $d$-based attribution
and random removal (mean $\pm$ across-seed SD, ten seeds per
condition).
The 0\% column is the baseline both arms share, so it is identical across rows by construction. Removing the same number of training subjects at random leaves $d$ near its baseline at every level.
Type~2 diabetes, for instance, holds $0.603$ from 5\% through 25\%
against a baseline of $0.601$, and is still at $0.595$ after half the training pool is gone, so the attribution arm's rise is not what removing data does on its own. The SDs here are dominated by the cohort draw, which both arms share within a seed and cancels in the paired change. See Figure~\ref{fig:removal} for that paired change, the quantity to read for significance.}
\label{tab:main}
\small
\resizebox{\textwidth}{!}{%
\begin{tabular}{llrrrrr}
\toprule
Condition & Arm & 0\% & 5\% & 10\% & 25\% & 50\% \\
\midrule
Multiple sclerosis & attribution & $0.671 \pm 0.148$ & $\boldsymbol{0.818 \pm 0.150}$ & $\boldsymbol{0.871 \pm 0.148}$ & $\boldsymbol{0.934 \pm 0.141}$ & $\boldsymbol{0.969 \pm 0.137}$ \\
 & random & & $0.669 \pm 0.148$ & $0.665 \pm 0.148$ & $0.660 \pm 0.146$ & $0.652 \pm 0.145$ \\
\addlinespace
Cerebrovascular & attribution & $0.333 \pm 0.056$ & $\boldsymbol{0.371 \pm 0.058}$ & $\boldsymbol{0.384 \pm 0.059}$ & $\boldsymbol{0.397 \pm 0.058}$ & $\boldsymbol{0.391 \pm 0.061}$ \\
 & random & & $0.333 \pm 0.056$ & $0.333 \pm 0.056$ & $0.333 \pm 0.056$ & $0.329 \pm 0.056$ \\
\addlinespace
Type~2 diabetes & attribution & $0.601 \pm 0.050$ & $\boldsymbol{1.269 \pm 0.041}$ & $\boldsymbol{1.439 \pm 0.033}$ & $\boldsymbol{1.601 \pm 0.029}$ & $\boldsymbol{1.659 \pm 0.028}$ \\
 & random & & $0.603 \pm 0.051$ & $0.603 \pm 0.051$ & $0.603 \pm 0.049$ & $0.595 \pm 0.053$ \\
\addlinespace
Chronic kidney dis. & attribution & $0.359 \pm 0.051$ & $\boldsymbol{0.761 \pm 0.053}$ & $\boldsymbol{0.919 \pm 0.048}$ & $\boldsymbol{1.079 \pm 0.039}$ & $\boldsymbol{1.127 \pm 0.032}$ \\
 & random & & $0.358 \pm 0.051$ & $0.358 \pm 0.051$ & $0.359 \pm 0.050$ & $0.356 \pm 0.050$ \\
\bottomrule
\end{tabular}%
}
\end{table}

\begin{figure}[h]
\centering
\includegraphics[width=1\linewidth]{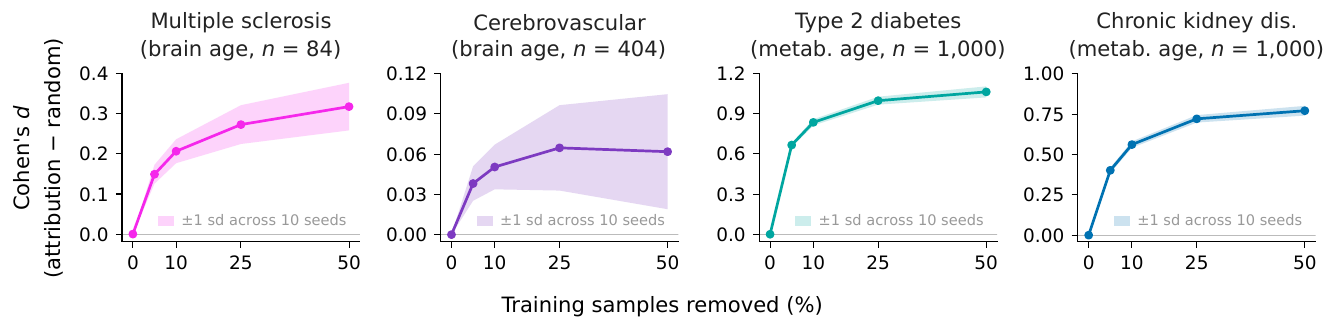}
\caption{Influence-guided removal raises the evaluation-half case-control effect size at every removal level in all four conditions. Per condition, we report the paired $\Delta d$ between Cohen's $d$-based attribution and random removal, averaged across seeds. Pairing removes cohort-draw noise shared by both arms. Coloured curves show attribution-based results, one colour per condition, with shaded $\pm 1$~SD across ten seeds. The grey zero line denotes random removal by construction.}
\label{fig:removal}
\end{figure}

Across four conditions, two modalities, two attribution targets, and all
removal levels, the paired change is positive across every seed-level cell we tested (Figure~\ref{fig:removal}, Appendix Figure~\ref{fig:app:meanbag}). Random removal is centred on zero at every level, with sign counts consistent with chance, including 50\%
removal, where the model trains on half the data. The effect comes from the
ranking, not from the change in training-set size. 

Beyond the size of the gain, which group drives the separation matters clinically. A useful biomarker keeps controls near an age gap $g$ of zero and shows a large gap only for cases. Before removal, healthy controls have a mean gap within $0.25$~y of zero. Their MAE is $3.2$ and $3.4$~y for brain age and $9.9$ and $10.5$~y for metabolomic age (Appendix Table~\ref{tab:mae_gap_d}). Random removal leaves both unchanged, while attribution-based removal widens the case gap and keeps the control gap within $1.2$~y of zero in every condition (Figure~\ref{fig:intro}, Appendix Figure~\ref{fig:age_gap_all}). Healthy accuracy holds or improves. For brain age, the controls' MAE is unchanged, and for metabolomic age it improves by about one year. This is consistent with the contamination hypothesis, as removing the flagged samples brings the model closer to healthy ageing while making it more sensitive to disease.

Paired gains are larger for the metabolomic conditions than for the brain conditions: $+0.84$ and $+0.56$ at 10\% removal against $+0.21$ and $+0.05$. This ordering would be expected under reference-cohort contamination if
metabolic dysfunction is more common in nominally healthy cohorts than subclinical brain disease.

\subsection{Choice of attribution target}
\label{sec:crossover}
Across the two attribution targets, Cohen's $d$ and the mean age gap (Section~\ref{sec:setup}), the gains in Table~\ref{tab:main} shift by at most 0.08 in $d$, small
against the gains themselves, and no sign or ordering changes. Which target
wins depends on the selection-half size (Table~\ref{tab:meangap_diff}). For
type~2 diabetes and chronic kidney disease, 
1{,}000 cases per group, Cohen's $d$ wins by 0.007 to 0.033 at every
removal level, small in absolute terms but decisive given the paired
standard deviation, four to ten times tighter than the marginal spread. For cerebrovascular disease
(404 cases) and multiple sclerosis (84 cases), the mean age gap wins instead, by 0.004 to 0.006 and by
0.043 to 0.079. The mean-gap surrogate drops the pooled-SD denominator, so the two targets differ only in the gradient through $s_{\text{pooled}}$. That term is estimated from the deviations of the
selection-half subjects around their group means. At 1{,}000 cases it is
informative about how a training sample shapes within-group spread. At 404
and 84 cases it is dominated by sampling noise of the selection-half draw
instead, so the fitted component overfits that particular sample and does
not transfer to the evaluation half. Consequently, the choice of attribution target is a bias-variance tradeoff, with $d$ being the estimand but noisier at a small selection half, and the mean age gap
being mis-specified but stable. When the selection half is small, prefer the
surrogate. The same bias-variance tradeoff can arise for other finite-sample
evaluation targets, including AUROC and calibration slopes.

\subsection{Who gets flagged}
\label{sec:pheno}

\begin{figure}[h]
\centering
\includegraphics[width=0.9\linewidth]{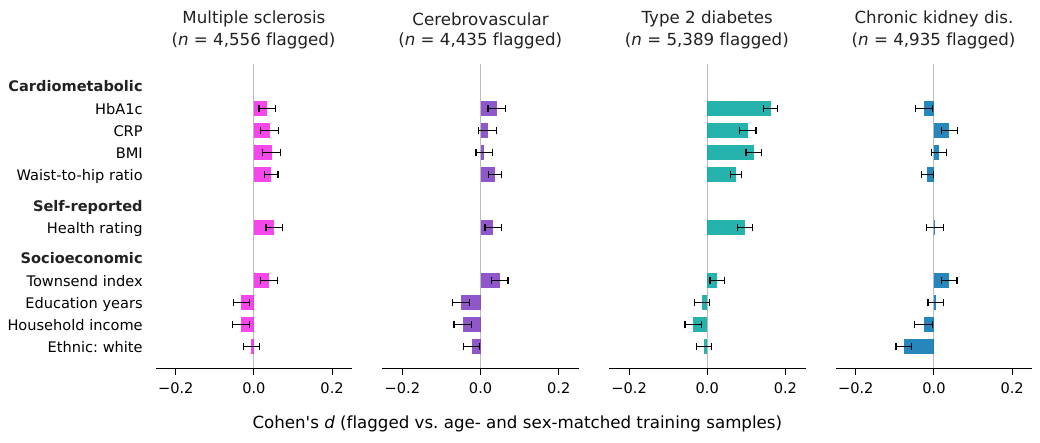}
\caption{Subjects the method flags differ from matched unflagged training subjects on
cardiometabolic health, self-reported health, and socioeconomic and
demographic measures. Bars are Cohen's $d$ between the flagged training samples and age- and sex-matched training samples drawn from the unflagged remainder of the same training pool. Error bars are $\pm 1$ bootstrap standard error. Age and sex are the matching variables and are not shown, since they are equal by
construction. Flagged subjects are the top 10\% by percentile rank averaged across the ten seeds. Flagging, matching, and the bootstrap follow Appendix~\ref{app:pheno} ($n = 4{,}556$, $4{,}435$, $5{,}389$, and $4{,}935$ flagged, given in each panel title).
Effects are largest in type~2 diabetes, the condition the method helps most: HbA1c $+0.162$, BMI $+0.119$, CRP $+0.104$, health rating $+0.097$. Chronic kidney disease breaks this pattern. Its cardiometabolic bars sit near zero, while ethnicity shows the largest separation in that panel ($d = -0.077$), larger than any cardiometabolic marker in the same condition (CRP, $+0.039$).}
\label{fig:phenotype}
\end{figure}

In three of the four conditions, the top 10\% of flagged samples have higher BMI, waist-to-hip ratio, C-reactive protein (CRP), and HbA1c, and report worse self-reported health than matched unflagged subjects from the same training pool (Figure~\ref{fig:phenotype}). Age and sex are balanced by construction. The pattern
is strongest for type~2 diabetes, an HbA1c separation of $d \approx 0.16$, more than three times the separation in any other run. This is the expected marker for the condition, recovered without the method seeing it, and serves as a positive control. It is absent for chronic kidney disease, where these same markers are flat or reversed.

Figure~\ref{fig:phenotype} also plots the socioeconomic and ethnicity block directly. Multiple sclerosis, cerebrovascular disease, and
type~2 diabetes show no substantial demographic separation there, at most
$d = -0.050$ for education years under cerebrovascular disease. Chronic kidney disease does not follow this pattern. It shows an ethnicity gap ($d = -0.077$), larger than any cardiometabolic marker. This separation raises a fairness question, which Section~\ref{sec:limitations} discusses.
The signature is shared across three conditions and strongest where the gain is largest. Chronic kidney disease is the exception.

\section{Limitations}
\label{sec:limitations}

The model class is ridge regression, where
influence is in closed form. We have not tested whether these rankings transfer to nonlinear age models. Such transfer is known to be delicate \citep{basu2021}, though ridge and nonlinear models perform comparably for age prediction \citep{perf_reserves}. We compute a batch-wide
ranking once. An iterative variant could behave differently if
contamination has cascade structure. We do not compare against
conventional, loss-based attribution baselines, since a loss-based ranking does not target the group-level effect size we evaluate. Deleting individuals is the crudest use of the ranking. The intended use, deriving interpretable inclusion criteria from the flagged profiles, is future work.

Current practice does not exclude one condition at a time. Normative age models are usually trained once across all conditions, with a single reference cohort excluding every condition under study, since these conditions share an elevated biological age signal. We instead train one model
per condition, excluding only that condition from its own reference
cohort. Whether our reported effect-size gains grow or shrink under
such a shared, more strictly filtered normative cohort remains open.
Results are from one biobank, and the two metabolomic conditions used
a capped control pool whose effect on absolute baselines we have not
yet isolated. The paired $\dd$ results do not depend on the baselines,
but the absolute effect sizes do.

Reference-cohort contamination is our working hypothesis for why removal helps. The phenotype and accuracy results are consistent with it but do not establish it. The phenotype panel (Section~\ref{sec:pheno}) describes the flagged group, not why the method flagged them. Ranking is based on influence on the age-gap effect size, not on these variables. For chronic kidney disease, the strongest separation is on ethnicity, a documented but untested correlate of disease risk.
Establishing a causal link is outside the scope of this analysis, so
we cannot tell whether this reflects a clinical signal or a fairness
concern. 

\section{Recommendations}
Three reporting rules follow.
First, pair every attribution arm with a
count-matched random arm under identical seeds, otherwise removal confounds attribution with training-set size. Second, report the paired change between the attribution and random arms
rather than absolute values or ratios of means, which inherit and can
amplify cohort-draw variance. Third, report the flagged-sample phenotype table
alongside the removal curves.

\section{Conclusion} 
Retargeting attribution from the training loss
to disease-related effect size identifies training samples missed by
diagnosis-based exclusions. Across four diseases and two modalities,
removing these samples consistently increases held-out effect size
while random removal does not. The approach should
generalise beyond Cohen's $d$ to any evaluation statistic computed on
held-out data, wherever the target that matters is not the one
attribution conventionally optimises.

\begin{ack}
We thank the UKBB participants for their voluntary commitment and the
UKBB team for their work in collecting, processing, and disseminating
these data for analysis. This research was conducted using the UKBB
Resource under project-ID 33073. The project was funded by the Deutsche
Forschungsgemeinschaft (DFG, German Research Foundation) under
project-IDs 565356445 and 586414057 to MAS.
\end{ack}

\newpage
\bibliographystyle{plainnat}
\bibliography{references}

\newpage
\appendix

\newpage

\section{AI disclosure}
\label{app:llm}
We used LLM assistants (e.g.\ ChatGPT, Claude) as coding co-pilots for
experimentation and plotting code, and as writing tools for language
editing, condensing drafts, and checking citation coverage. All
scientific ideas, experimental design, claims, and cited references
originate from and were verified by the authors, who reviewed and take
full responsibility for the final content.

\newpage
\section{The age-gap biomarker}
\label{app:agegap}

Readers coming from the attribution literature may not be familiar with
how an age-gap biomarker is built or why it is used clinically. This
section gives that context briefly.

An age-gap biomarker predicts a person's age from a biological
measurement, typically brain imaging \citep{cole2017, estimation_smith} or a blood panel \citep{metabolomic}, using a model
trained only on people presumed healthy. This is an instance of
normative modelling \citep{brain_age_paradigm_franke}. A model captures the typical ageing trajectory of a reference cohort, and an individual's deviation from that trajectory serves as a biomarker. Applying the model to a new subject $i$ with true age $y_i$
and predicted age $\hat y_i$ gives the age gap $g_i = \hat y_i - y_i$: a
value near zero means the subject's biology matches the healthy
trajectory, while a large positive value means the model reads their
biology as older than their years \citep{cole2017}. For example, a
subject with a true age of 50 and a predicted age of 55 has an age gap $g$
of five years. Section~\ref{sec:method} formalises this further with a
bias correction $(a,b)$ fit on training predictions, giving
$g_i = (\hat y_i - b)/a - y_i$.

The biomarker becomes clinically useful once this gap is compared
between people with a disease and matched healthy controls. Writing
$\bar g_{\text{case}}$ and $\bar g_{\text{ctrl}}$ for the mean gap in
each group, a disease that accelerates biological ageing shows up as
$\bar g_{\text{case}} > \bar g_{\text{ctrl}}$, and a wider gap has been
linked to more severe symptoms and faster progression across a range of
conditions \citep{brain_age_paradigm_cole2}. The size of this
separation, normalised by the pooled standard deviation as Cohen's $d$, is the biomarker's disease-related effect size. This effect size is evaluated on people the model never trained on, not on the model's training accuracy.

Two consequences of this setup motivate our method. First,
like any normative modelling framework, this construction depends on the
reference cohort being free from the pathological patterns the
model is meant to be sensitive to \citep{rutherford2022normative}. Any
undiagnosed or subclinical disease hiding in the presumed-healthy
training pool distorts what the model treats as a normal ageing
trajectory, which is the contamination problem Section~\ref{sec:method}
addresses. Second, since the effect size is measured downstream
on held-out cases and controls rather than on training loss, a model that fits its training data less tightly can still produce a gap with a larger effect size \citep{schulz2025}. This is why we attribute the
downstream effect size $F$ directly, rather than the training loss the
model was fit to minimise.

Although brain imaging is the most established input for this
biomarker, the same construction applies to any modality carrying a
biological age signal. This paper evaluates two: brain age from T1
imaging phenotypes and metabolomic age from NMR blood metabolomics,
both fit and evaluated identically.

\newpage
\section{Experimental setup}
\label{app:setup}
In this section, we record the compute and every pipeline-specific design
choice needed to reproduce the reported results. Standard components, such
as ridge regression, the leave-one-out retraining, and matching, follow their
usual textbook form, so we do not re-derive them here.

\subsection{Cohorts and training pool}
\label{app:tables}
Four conditions across two UK Biobank modalities: brain age from 1{,}440 T1
imaging features (45{,}755 subjects, visit 2) and metabolomic age from 251 NMR
features (98{,}349 subjects, baseline visit). Status comes from ICD-10 codes
relative to the assay visit. We reserve prevalent cases for influence computation and evaluation only, exclude incident (post-visit) diagnoses entirely, and treat subjects with no record of the diagnosis as controls.

\begin{table}[h]
\centering
\caption{Cohort sizes by condition.}
\label{tab:cohorts}
\footnotesize
\begin{tabular}{lrrr}
\toprule
Condition & Prevalent & Incident & Control \\
\midrule
Multiple sclerosis (\texttt{G35})   & 168     & 12      & 45{,}575 \\
Cerebrovascular (\texttt{I60--I69}) & 808     & 590     & 44{,}357 \\
Type 2 diabetes (\texttt{E11})      & 7{,}567 & 18{,}478 & 72{,}304 \\
Chronic kidney dis. (\texttt{N18})  & 3{,}419 & 14{,}084 & 80{,}846 \\
\bottomrule
\end{tabular}
\end{table}

We exclude all prevalent and incident subjects for the condition under study
from the training pool and cap it at 40{,}000 by a \emph{per-seed}
subsample. Each seed draws its own 40{,}000 from the eligible pool, so a
subject appears in a variable number of seeds. 
In the T1 cohorts, nearly every eligible subject is drawn by at least five seeds, while in the NMR cohorts only 75\% (\texttt{E11}) and 61\% (\texttt{N18}) are. The seed-averaged ranking in Section~\ref{app:pheno} therefore only includes subjects drawn in at least five seeds, so that every averaged rank rests on enough draws.
Beyond this cap, the pipeline applies a 5\% feature-missingness filter (training set) and
a 90\% subject-coverage filter. Neither filter removes anything in either cohort.

\subsection{Matching and evaluation split}
\label{app:matching}
We cap cases at 2{,}000 and match them 1:1 to controls by nearest-neighbour
Mahalanobis distance on age and sex (\matchpkg{} \citep{cohortmatch}, ATT estimand,
average treatment effect on the treated, exact sex match, 2-year age
caliper), reducing the mean absolute standardised mean difference from
0.38--0.54 across all condition-seed matchings to 0.000 in every one
of them. We split matched pairs 50/50 by pair into a selection half, used
only to compute attribution scores, and an evaluation half, serving as the test set, used only to report effect sizes.

\subsection{Model and attribution}
\label{app:model}
The age model is ridge regression (scikit-learn) on the modality's features
plus sex. We fit mean imputation and standardisation on the training set.
We choose the penalty $\alpha$ once per seed by cross-validation on the untouched pool, and freeze it for every refit. The
bias-corrected age gap is $g = (\hat y - b)/a - y$. We fit $(a,b)$ by
least squares of training predictions on training age
(Section~\ref{sec:method}).

We score each training subject under four configurations: Cohen's $d$ of $g$ and the
unnormalised mean age gap $\bar g_{\text{case}} - \bar g_{\text{ctrl}}$,
each by closed-form influence function (\infpkg{}) and by exact
leave-one-out retraining, the standard closed-form ridge update, with the
intercept left unpenalised to match the model fit.

\subsection{Removal experiment and effect sizes}
\label{app:removal}
We remove training subjects in descending score order at 5, 10, 25, and
50\% cuts, refit the model at frozen $\alpha$, and recompute Cohen's $d$ on
the evaluation half. 

For each removal level, we compute the paired change $\dd$, the
attribution arm's Cohen's $d$ minus the random arm's, within each
seed, and report its mean and standard deviation across seeds.
This cancels the cohort-draw variance common to every arm within
a seed. Its error bar is the across-seed SD (Figure~\ref{fig:removal}).

\subsection{Phenotypic characterisation}
\label{app:pheno}
We flag the top 10\% most influential training subjects by percentile rank
averaged across the ten seeds, match them 1:1 against the remaining
training subjects (nearest neighbour on age, exact on sex, caliper
2~years) and compare the two groups on the variables shown in
Figure~\ref{fig:phenotype} by Cohen's~$d$ with $\pm 1$ bootstrap standard
error ($B = 1{,}000$). Each seed ranks only its own 40{,}000-subject
draw, so we convert to percentile rank before averaging and require
subjects appearing in the ranking in at least five seeds. The pooled ranking's support is the set of subjects drawn in at least five seeds, 45{,}557, 44{,}352, 53{,}890,
and 49{,}349 subjects by condition, so the flagged top 10\% number
4{,}556, 4{,}435, 5{,}389, and 4{,}935 rather than a round 4{,}000.
Because the design is 1:1 matched, the bootstrap resamples \emph{pairs}.
One index vector draws both groups, so a flagged subject and their matched unflagged subject travel together and the interval conditions on the matching that
produced it. We carry age and sex through as a matching check. This design
makes both members of a pair identical on these variables, so $d = 0.000$ with a standard
error of exactly zero in every panel.

\subsection{Seeds}
\label{app:seeds}
A single seed controls every stochastic step (subsampling, matching, the
training draw, the selection/evaluation split, random removal, and the
bootstrap). For the removal experiment, error bars are the across-seed spread
of the paired change.
Figure~\ref{fig:phenotype} uses the aggregate-ranking bootstrap described in section~\ref{app:pheno}.

\subsection{Computational details}
\label{app:compute}
We run all experiments on CPU-only AMD EPYC nodes, no GPU, requesting 8
cores and 32\,GB per condition-seed task, with BLAS threaded across the 8
cores and no other parallelism. Peak memory is under 6\,GB per task.

Per-task wall-clock ranges from 43\,s to 333\,s, driven by predictor count
(1{,}441 for T1 imaging against 252 for NMR) rather than sample size. The
NMR tasks run in 43--72\,s on the larger cohorts, the T1 tasks in
114--333\,s. The full set of 40 tasks (4 conditions $\times$ 10 seeds)
totals roughly 1.6 hours of compute, about 12.5 core-hours. That figure covers only the final experiments reported in this paper. Total project compute is higher,
since we ran the pipeline repeatedly during development and debugging before
reaching this configuration.

Software: Python 3.12, scikit-learn 1.8, plus \matchpkg{} and
\infpkg{}, both released under the MIT License.

\newpage
\section{Removal effect on mean age gap and prediction error}
\label{app:years}

\begin{table}[h]
\centering
\caption{Evaluation-half absolute prediction error (MAE) and mean age gap, in years, for healthy controls and their matched cases under Cohen's $d$ attribution and random removal at $10\%$ (mean $\pm$ across-seed SD, ten seeds per condition). Random removal leaves both groups at baseline. Attribution raises the cases' error and gap but leaves the controls' almost unchanged, apart from the two metabolomic conditions, where their error drops by about one year and their gap by 0.5 to 0.9 years.}
\label{tab:mae_gap_d}
\small
\begin{tabular}{lllrrr}
\toprule
Condition & Metric & Group & 0\% & random & attribution \\
\midrule
Multiple sclerosis & MAE & controls & $3.16 \pm 0.21$ & $3.17 \pm 0.22$ & $3.16 \pm 0.23$ \\
 & & cases & $5.02 \pm 0.35$ & $5.02 \pm 0.35$ & $6.64 \pm 0.48$ \\
 & Mean age gap & controls & $-0.11 \pm 0.41$ & $-0.10 \pm 0.42$ & $-0.01 \pm 0.42$ \\
 & & cases & $+3.33 \pm 0.52$ & $+3.31 \pm 0.51$ & $+5.24 \pm 0.65$ \\
\addlinespace
Cerebrovascular & MAE & controls & $3.35 \pm 0.11$ & $3.36 \pm 0.11$ & $3.35 \pm 0.10$ \\
 & & cases & $3.92 \pm 0.08$ & $3.93 \pm 0.09$ & $4.00 \pm 0.11$ \\
 & Mean age gap & controls & $-0.02 \pm 0.18$ & $-0.03 \pm 0.18$ & $-0.20 \pm 0.19$ \\
 & & cases & $+1.47 \pm 0.12$ & $+1.46 \pm 0.12$ & $+1.55 \pm 0.17$ \\
\addlinespace
Type 2 diabetes & MAE & controls & $9.93 \pm 0.20$ & $9.93 \pm 0.20$ & $9.01 \pm 0.13$ \\
 & & cases & $13.81 \pm 0.50$ & $13.82 \pm 0.54$ & $27.14 \pm 0.68$ \\
 & Mean age gap & controls & $+0.18 \pm 0.30$ & $+0.18 \pm 0.29$ & $-0.36 \pm 0.36$ \\
 & & cases & $+8.68 \pm 0.63$ & $+8.71 \pm 0.66$ & $+25.38 \pm 0.62$ \\
\addlinespace
Chronic kidney dis. & MAE & controls & $10.52 \pm 0.19$ & $10.52 \pm 0.20$ & $9.36 \pm 0.19$ \\
 & & cases & $12.02 \pm 0.32$ & $12.02 \pm 0.33$ & $16.54 \pm 0.40$ \\
 & Mean age gap & controls & $-0.24 \pm 0.41$ & $-0.25 \pm 0.41$ & $-1.13 \pm 0.38$ \\
 & & cases & $+4.84 \pm 0.40$ & $+4.81 \pm 0.40$ & $+13.12 \pm 0.43$ \\
\bottomrule
\end{tabular}
\end{table}

\begin{figure}[h]
\centering
\includegraphics[width=0.85\linewidth]{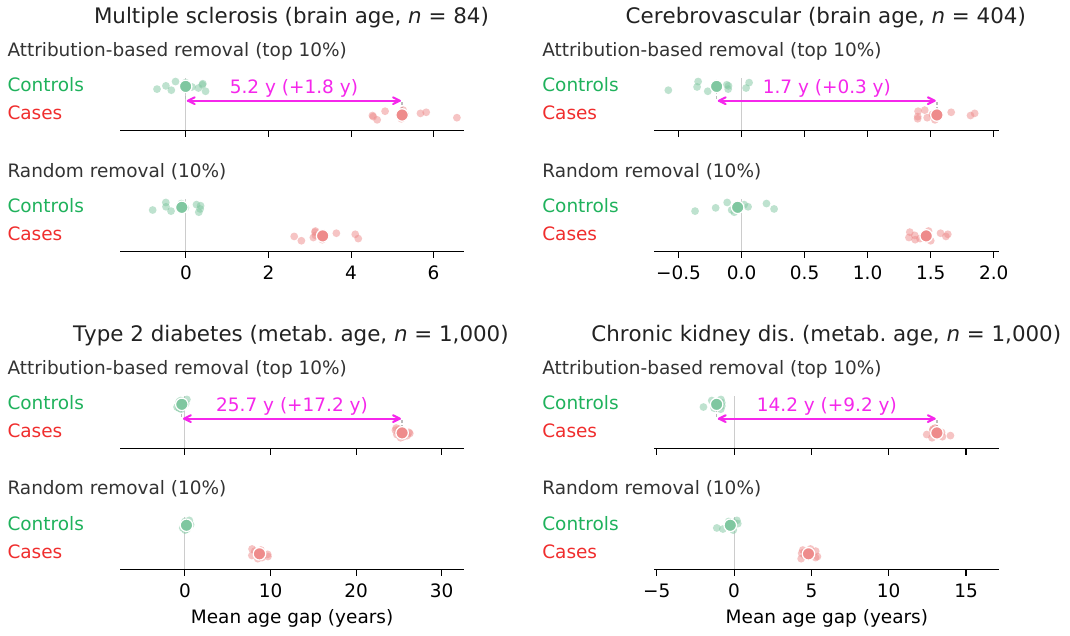}
\caption{Mean age gap shift under Cohen's $d$ attribution at $10\%$ removal, as in Figure~\ref{fig:intro}(c), for all four conditions. Upper row of each block: attribution-based removal. Lower row: random removal. Large markers show the across-seed mean (green: matched controls, red: cases) and small lighter dots the single-seed means, in place of error bars. The arrow gives the case-control gap under attribution and, in brackets, its paired gain over random removal. See Table~\ref{tab:mae_gap_d} for the values. Brain age and metabolomic age are different modalities, so their scales are not comparable in years.}
\label{fig:age_gap_all}
\end{figure}

\newpage
\section{Influence distribution}
\label{app:infdist}

\begin{figure}[h]
\centering
\includegraphics[width=\linewidth]{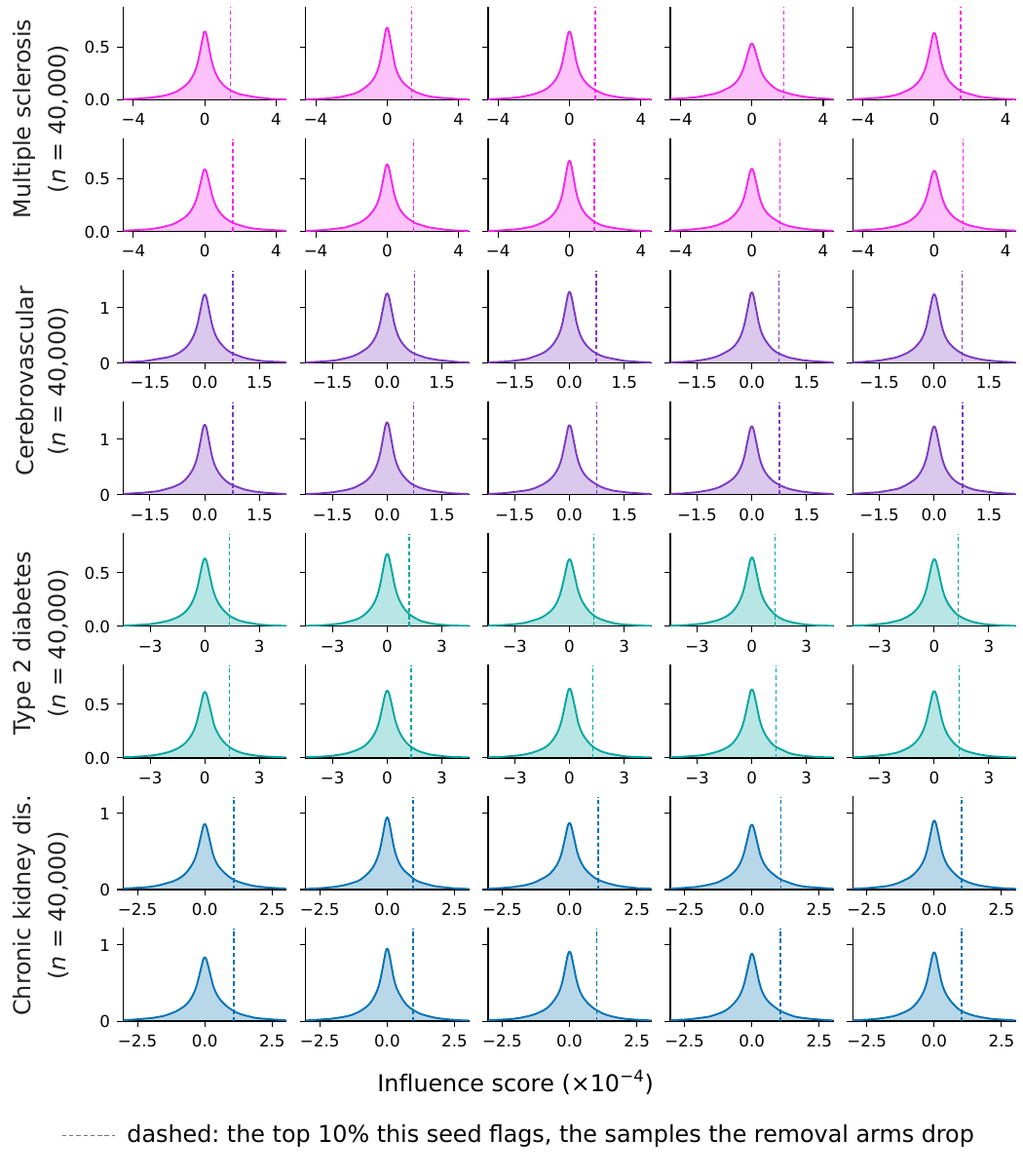}
\caption{Distribution of Cohen's $d$ influence scores across training
samples, shown per condition (rows) and per seed (columns). Every panel
shows the same sharply peaked, heavy-tailed shape, consistent with
influence functions generally. The dashed line marks the
10\% removal cutoff.}
\label{fig:app:infdist:cohensd}
\end{figure}

\newpage
\section{Mean age gap attribution}
\label{app:meanbag}
\raggedbottom

\begin{table}[H]
\centering
\caption{Evaluation-half Cohen's $d$ under mean age gap attribution
and random removal (mean $\pm$ across-seed SD, ten seeds per
condition). The SDs here are dominated by the cohort draw, which both arms share within a seed and cancels in the paired change. See Figure~\ref{fig:app:meanbag} for that paired change, the quantity to read for significance.}
\label{tab:meangap}
\small
\resizebox{\textwidth}{!}{%
\begin{tabular}{llrrrrr}
\toprule
Condition & Arm & 0\% & 5\% & 10\% & 25\% & 50\% \\
\midrule
Multiple sclerosis & attribution & $0.671 \pm 0.148$ & $\boldsymbol{0.861 \pm 0.139}$ & $\boldsymbol{0.926 \pm 0.131}$ & $\boldsymbol{1.005 \pm 0.122}$ & $\boldsymbol{1.049 \pm 0.122}$ \\
 & random & & $0.669 \pm 0.148$ & $0.665 \pm 0.148$ & $0.660 \pm 0.146$ & $0.652 \pm 0.145$ \\
\addlinespace
Cerebrovascular & attribution & $0.333 \pm 0.056$ & $\boldsymbol{0.375 \pm 0.056}$ & $\boldsymbol{0.390 \pm 0.056}$ & $\boldsymbol{0.403 \pm 0.055}$ & $\boldsymbol{0.397 \pm 0.060}$ \\
 & random & & $0.333 \pm 0.056$ & $0.333 \pm 0.056$ & $0.333 \pm 0.056$ & $0.329 \pm 0.056$ \\
\addlinespace
Type~2 diabetes & attribution & $0.601 \pm 0.050$ & $\boldsymbol{1.251 \pm 0.041}$ & $\boldsymbol{1.414 \pm 0.032}$ & $\boldsymbol{1.568 \pm 0.029}$ & $\boldsymbol{1.626 \pm 0.028}$ \\
 & random & & $0.603 \pm 0.051$ & $0.603 \pm 0.051$ & $0.603 \pm 0.049$ & $0.595 \pm 0.053$ \\
\addlinespace
Chronic kidney dis. & attribution & $0.359 \pm 0.051$ & $\boldsymbol{0.753 \pm 0.052}$ & $\boldsymbol{0.903 \pm 0.049}$ & $\boldsymbol{1.058 \pm 0.042}$ & $\boldsymbol{1.109 \pm 0.035}$ \\
 & random & & $0.359 \pm 0.051$ & $0.358 \pm 0.051$ & $0.359 \pm 0.050$ & $0.356 \pm 0.050$ \\
\bottomrule
\end{tabular}%
}
\end{table}

\begin{figure}[H]
\centering
\includegraphics[width=1\linewidth]{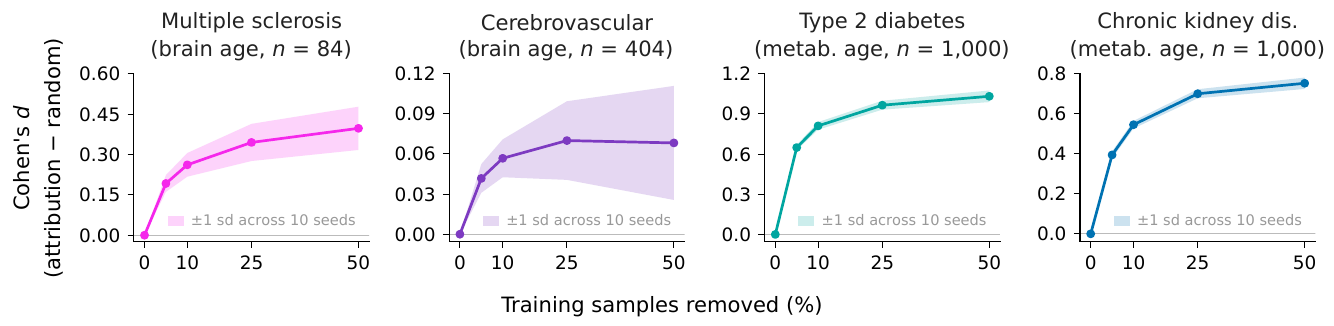}
\caption{Influence-guided removal raises the evaluation-half case-control effect size at every removal level, in all four conditions.
Per condition, we report the paired $\Delta d$ between mean age gap attribution and random-removal results, averaged across seeds. Pairing removes cohort-draw noise shared by both arms. Coloured curves represent attribution-based results, one
colour per condition, with shaded $\pm 1$~SD across ten seeds. The grey line at
zero represents random removal by construction.}
\label{fig:app:meanbag}
\end{figure}

\begin{table}[H]
\centering
\caption{Difference between the two attribution targets, mean age gap minus
Cohen's $d$, in evaluation-half Cohen's $d$. The comparison is paired
within seed. Both targets share that seed's matched set, selection/evaluation
split, frozen $\alpha$ and baseline, so the cohort-draw variance common to both
differences out. The SDs here are consequently four to ten times smaller than
the marginal SDs in Tables~\ref{tab:main} and~\ref{tab:meangap}, and the two
should not be compared directly: a difference of $0.03$ is decisive here while
each individual curve carries an SD of $0.03$--$0.15$. Positive favours the mean age
gap attribution.}
\label{tab:meangap_diff}
\small
\begin{tabular}{lrrrr}
\toprule
Condition & 5\% & 10\% & 25\% & 50\% \\
\midrule
Multiple sclerosis & $+0.043 \pm 0.018$ & $+0.055 \pm 0.026$ & $+0.072 \pm 0.034$ & $+0.079 \pm 0.033$ \\
Cerebrovascular & $+0.004 \pm 0.004$ & $+0.006 \pm 0.007$ & $+0.005 \pm 0.008$ & $+0.006 \pm 0.008$ \\
Type 2 diabetes & $-0.018 \pm 0.006$ & $-0.026 \pm 0.008$ & $-0.033 \pm 0.007$ & $-0.033 \pm 0.008$ \\
Chronic kidney dis. & $-0.007 \pm 0.005$ & $-0.016 \pm 0.002$ & $-0.021 \pm 0.004$ & $-0.018 \pm 0.005$ \\
\bottomrule
\end{tabular}
\end{table}

\begin{table}[H]
\centering
\caption{Evaluation-half absolute prediction error (MAE) and mean age gap, in years, for healthy controls and their matched cases under mean age gap attribution and random removal at $10\%$ (mean $\pm$ across-seed SD, ten seeds per condition). Random removal leaves both groups at baseline. Attribution raises the cases' error and gap but leaves the controls' almost unchanged, apart from the two metabolomic conditions, where their error drops by about one year and their gap by 0.6 to 0.9 years.}
\label{tab:mae_gap_meangap}
\small
\begin{tabular}{lllrrr}
\toprule
Condition & Metric & Group & 0\% & random & attribution \\
\midrule
Multiple sclerosis & MAE & controls & $3.16 \pm 0.21$ & $3.17 \pm 0.22$ & $3.16 \pm 0.24$ \\
 & & cases & $5.02 \pm 0.35$ & $5.02 \pm 0.35$ & $7.21 \pm 0.40$ \\
 & Mean age gap & controls & $-0.11 \pm 0.41$ & $-0.10 \pm 0.42$ & $0.00 \pm 0.39$ \\
 & & cases & $+3.33 \pm 0.52$ & $+3.31 \pm 0.51$ & $+5.90 \pm 0.57$ \\
\addlinespace
Cerebrovascular & MAE & controls & $3.35 \pm 0.11$ & $3.36 \pm 0.11$ & $3.37 \pm 0.11$ \\
 & & cases & $3.92 \pm 0.08$ & $3.93 \pm 0.09$ & $4.07 \pm 0.11$ \\
 & Mean age gap & controls & $-0.02 \pm 0.18$ & $-0.03 \pm 0.18$ & $-0.16 \pm 0.20$ \\
 & & cases & $+1.47 \pm 0.12$ & $+1.46 \pm 0.12$ & $+1.64 \pm 0.15$ \\
\addlinespace
Type 2 diabetes & MAE & controls & $9.93 \pm 0.20$ & $9.93 \pm 0.20$ & $8.96 \pm 0.13$ \\
 & & cases & $13.81 \pm 0.50$ & $13.82 \pm 0.54$ & $26.49 \pm 0.70$ \\
 & Mean age gap & controls & $+0.18 \pm 0.30$ & $+0.18 \pm 0.29$ & $-0.38 \pm 0.35$ \\
 & & cases & $+8.68 \pm 0.63$ & $+8.71 \pm 0.66$ & $+24.63 \pm 0.61$ \\
\addlinespace
Chronic kidney dis. & MAE & controls & $10.52 \pm 0.19$ & $10.52 \pm 0.20$ & $9.29 \pm 0.18$ \\
 & & cases & $12.02 \pm 0.32$ & $12.02 \pm 0.33$ & $16.15 \pm 0.36$ \\
 & Mean age gap & controls & $-0.24 \pm 0.41$ & $-0.25 \pm 0.41$ & $-1.16 \pm 0.36$ \\
 & & cases & $+4.84 \pm 0.40$ & $+4.81 \pm 0.40$ & $+12.64 \pm 0.42$ \\
\bottomrule
\end{tabular}
\end{table}

\clearpage
\flushbottom

\end{document}